\documentclass[runningheads]{llncs}

\usepackage{graphicx}
\usepackage{amsmath,amssymb}
\usepackage{mathabx}

\usepackage{graphicx}
\usepackage{xcolor}
\usepackage[outercaption]{sidecap}
\usepackage[font=scriptsize]{caption}
\usepackage{subcaption}
\usepackage{tabu}
\usepackage{wrapfig}
\usepackage{bbm}
\usepackage{algorithm}
\usepackage[noend]{algpseudocode}
\usepackage[utf8]{inputenc}
\usepackage[T1]{fontenc}   
\usepackage{hyperref}      
\usepackage{url}           
\usepackage{booktabs}      
\usepackage{amsfonts}      
\usepackage{nicefrac}      
\usepackage{microtype}     
\usepackage{xcolor} 
\usepackage{authblk}

\begin{document}
\title{Continual Model-based Reinforcement Learning for Data Efficient Wireless Network Optimisation}
\titlerunning{~}



\author{Cengis Hasan* \and 
        Alexandros Agapitos* \and 
        David Lynch* \and 
        Alberto Castagna \and 
        Giorgio Cruciata \and 
        Hao Wang \and 
        Aleksandar Milenovic} 
\authorrunning{C. Hasan et al.}        
\institute{Huawei Ireland Research Center \\
\email{\texttt{linyun.wanghao@huawei.com}} \\
*Authors with equal contribution}

\tocauthor{Cengis~Hasan}
\tocauthor{Alexandros~Agapitos}
\tocauthor{David~Lynch}
\tocauthor{Alberto~Castagna}
\tocauthor{Giorgio~Cruciata}
\tocauthor{Hao~Wang}
\tocauthor{Aleksandar~Milenovic}

\maketitle

\begin{abstract}

We present a method that addresses the pain point of long lead-time required to deploy cell-level parameter optimisation policies to new wireless network sites. Given a sequence of action spaces represented by overlapping subsets of cell-level configuration parameters provided by domain experts, we formulate throughput optimisation as Continual Reinforcement Learning of control policies. Simulation results suggest that the proposed system is able to shorten the end-to-end deployment lead-time by two-fold compared to a reinitialise-and-retrain baseline without any drop in optimisation gain. 



\end{abstract}

\section{Introduction} 


One of the major factors influencing the Quality of Experience (QoE) in wireless networks is the parameter configuration of the cells in a base-station. Incorrectly configured cells can interfere with neighbouring cells and degrade quality of service through inadequate coverage or over-utilisation. Traditionally, cell-level parameter configuration is realised before deployment, at which time the engineers have to anticipate diverse traffic conditions (i.e. user load), radio channel conditions, environment conditions (i.e. physical location and surroundings of a cell), and the complex relationship of QoE objective with other conflicting optimisation objectives in a wireless network (i.e. coverage, utilisation, power consumption). 

Existing approaches for configuring cell-level parameters are based on a discrete black-box optimisation problem formulation, which is often solved with surrogate-driven optimisation methods. Due to the combinatorial parameter space defined by the large number of cell-level configuration parameters (> 500 parameters), and the variable impact of the same set of parameters on the QoE in different deployment sites, selecting and  configuring parameters for a new system deployment is a stage-wise process. In this process, priorities are assigned to subsets of parameters based on domain knowledge, and these subsets are then configured in order of priority in a sequence of independent surrogate-driven optimisation runs. When QoS metrics reach target values, the parameter selection and configuration process is terminated and the parameter configuration is executed to each cell. Each independent optimisation run is composed of 7 days of data collection, followed by 10 days of iterative optimisation, amounting to a total of 17 days. Each consecutive parameter subset optimisation will add this amount of additional lead-time before the configured parameters are applied to each cell.

In this work, we developed a technology to reduce the overall lead-time required to optimise cell-level parameter configurations for a new wireless network site. We formulated the problem of stage-wise optimisation of parameter subsets as a \emph{Continual Reinforcement Learning} problem. This approach leverages forward transfer of knowledge between optimisation policies with overlapping subsets of actions in order to learn the ultimate policy in a data-efficient task-oriented fashion. Additionally it allows for a safe rollback to a policy of a previous subset, if objective KPIs do not improve, by avoiding catastrophic forgetting. Through a series of experiments, we demonstrate a two-fold reduction in deployment lead-time compared to a \emph{Reinitialise-and-Retrain} baseline. The main challenges addressed in this paper, with corresponding technical solutions, are summarised in Table \ref{tab:challenges}.

\begin{table}[htbp]
\scriptsize
\caption{Challenges and technical solutions.}
\setlength\extrarowheight{2.0pt} 
\begin{tabu} to 1\textwidth { | X[l] | X[l] | }
   \hline
   \textbf{Challenge} & \textbf{Technical Solution} \\
   \hline
    Data limitation in real wireless network trials, which are expensive and constrained to real-time. & Model-based RL. \\
   \hline    
    Time constraints on policy deployment lead-time to address a new candidate configuration parameter subset. & Continual RL. \\
    \hline
    High levels of noise in objective KPIs. & Probabilistic reward model explicitly accounting for aleatoric and epistemic uncertainty. \\
    \hline
    Inference time constrained to under 5 minutes for 20,000 cells. &  Learning-based solution as opposed to online planning at decision time. \\
    \hline
\end{tabu}
\label{tab:challenges}
\end{table}

The rest of the paper is organised as follows. Section~\ref{sec:relatedWork} summarises previous work on Reinforcement Learning (RL) for network parameter optimisation, and outlines the main classes of methods for Continual RL (CRL). Section~\ref{sec:method} presents the detailed description of the real-world dataset, the problem formulation, and the solution methods. Section~\ref{sec:evaluation} describes the experiment design, and Section~\ref{sec:results} analyses the experiment results. Finally, we conclude in Section~\ref{sec:conclusions} and propose future research.






    
    
    



\section{Related Work}\label{sec:relatedWork}
Previous research has addressed the problem of wireless network parameter optimisation using rule-based methods~\cite{conf/vtc/EckhardtKG11,DBLP:conf/pimrc/EisenblatterG08}, mathematical models~\cite{10.1109/TNET.2013.2294965}, or RL~\cite{DBLP:journals/tccn/BaleviA19,DBLP:conf/blackseecom/BotheMFI20,ericsson,Calabrese2018LearningRR,10.1007/s11277-016-3849-9,8792117,DBLP:journals/twc/ShafinCNHPZRL20,DBLP:conf/vtc/VannellaJP20}. A common characteristic of the aforementioned works is that the action space (adjustable network configuration parameters) is defined a priori at design time and is kept fixed. Scenarios where one wishes to dynamically extend the action space with additional configuration parameters can be solved in a sample-efficient way via a Continual RL problem formulation~\cite{Khetarpal2020TowardsCR}. The main classes of methods for achieving positive forward transfer while mitigating catastrophic forgetting in CRL tasks are summarised as follows:

\begin{itemize}
    \item \textbf{Parameter storage based} methods require multiple independently trained models to be stored for the different tasks. Catastrophic forgetting is overcome at the cost of storing parameters for each model. The space needed to store policies is linear in the number of tasks addressed. Unfortunately, this technique does not support knowledge transfer across tasks regardless of their potential similarities. 
    
    \item \textbf{Distillation} is the process of distilling a source model(s) to a target model \cite{rusu2015policy,schwarz2018progress,traore2019discorl,yin2017knowledge}. Distillation mitigates the need to store multiple models by compressing them into a single neural network. Hence, models trained on several source tasks can be distilled into a single network which captures shared experiences from all tasks. Despite reducing storage space, distillation still requires task specific layers to extrapolate features or to fit a task's action policy.

    \item \textbf{Rehearsal} consists of training using examples from both the current task and old examples from previously encountered tasks. This requires either an experience replay buffer to store old examples, or if storage space is limited, pseudo-rehearsal whereby examples are synthesized using a generative model~\cite{isele2018selective,li2017learning,rolnick2019experience}. When solving a novel task, models can retain performance achieved in previous tasks by continuously revisiting examples from same.

    \item \textbf{Regularization based methods} maintain a single model across multiple tasks. Catastrophic forgetting is mitigated by preventing parameters that are important for previous tasks from changing significantly when learning a new task~\cite{kirkpatrick2017overcoming,yoon2017lifelong}. When compared to \textit{Parameter Storage} methods, this approach reduces the storage space required since only one model is maintained. A drawback is that the initial model must have sufficient capacity to accommodate all future tasks, which are typically unknown upfront. Furthermore, it is not clear how to handle extension of the action space.

    \item \textbf{Modular architectures} facilitate CL by exploiting flexibility and compositionability of neural networks. Model capacity can be adjusted dynamically, making adaptation to unseen tasks possible without over-parameterising the model initially \cite{rusu2016progressive,schwarz2018progress,yoon2017lifelong}. It is sometimes helpful to decompose complex problems into easier sub-problems. In this case, neural modules can be combined and re-used across tasks \cite{devin2017learning,mendez2022modular}. For instance, a decision policy could be decomposed into one module to extract features and another to select actions. Modules are combined as required to achieve positive transfer, while catastrophic forgetting is prevented by storing lightweight modules trained on prior tasks.
    
\end{itemize}




In our optimisation problem, the order and the type of overlap across the sequence of cell-level configuration parameter subsets are defined by the domain expert. We selected \textit{Progress-and-Compress} (P\&C)~\cite{schwarz2018progress}, a method that has demonstrated good performance in CRL scenarios of evolving action space. P\&C is a hydrid between the methods of distillation, regularisation and modular architectures, and it can be decomposed into two steps. Firstly, the progress step, where the agent achieves positive transfer by expanding the architecture horizontally via lateral connections from an existing knowledge base (assimilates experience from all previous tasks) to a new active column (learns skills on the new task). Secondly, in the compress step, model distillation is used to distill the active column into the knowledge base without disrupting performance on previous tasks. As such, P\&C mitigates unbounded growth in learned policies, while ensuring transfer across similar tasks and preventing catastrophic forgetting.



\section{Methods}
\label{sec:method}
In this section, the dataset is described and the problem is formulated as a Markov Decision Process (MDP). Methods are introduced to achieve data efficient continual learning of network control policies using model-based RL.

\subsection{Description of the Dataset}\label{sec:prob_defn}
A dataset $D_{init}$ was collected over 5 days in a real 5G network containing 966 cells, by executing random actions to random cells at hourly intervals. We collected four different feature sets\footnote{The names of CPs, PCs, and EPs cannot be disclosed.}: cell-level configuration parameters (CPs) representing the actions, performance counters (PCs), engineering parameters (EPs), and spatio-temporal (ST) context. PCs include time-varying features (e.g. demand, channel quality, etc.) which are reported by cells hourly. EPs are fixed characteristics of a cell such as its tilt, antenna type, etc. The optimisation objective KPI is cell-level throughput measured in Gigabits per second (Gbps). Exploratory analysis revealed the following challenging properties of the dataset.

\textbf{High dimensional state and action spaces:} The composite action space consists of 19 CPs. Three different groups of CPs were adjusted during the data collection experiment. The three groups include: 4 CPs for power control on cells, 6 CPs for modulation code scheme selection (method used by a cell to encode digital data), and 9 CPs for rank selection (number of spatial streams used to transmit data from a cell to users). CP are categorical variables that take between 2 and 13 different values -- the number of unique action combinations is $\approx 4.6\times 10^{18}$. Raw network state is the union of PCs, EPs, and ST context. A total of 410 features constitute the network state. 


\textbf{Throughput (TP) time-series exhibits high levels of noise:} A seasonality-trend-noise decomposition of the time-series was performed for each cell. It was observed that $49.41\% \pm 10.04\%$ of the variance in TP is explained by seasonal and trend components. The remaining variance can be attributed to the noise component of the time-series.

The challenges and technical solutions are summarised in Table \ref{tab:challenges}. A tabular dataset was formed by aligning the four feature sets with respect to cell ID and timestamp. The aligned dataset contained $966\,\,[\mathrm{cells}]*5\,\,[\mathrm{days}]*24\,\,[\mathrm{hours}]$ examples. Rows with missing values due to energy saving enabled on some cells in early morning hours were removed, resulting in a total of 103,648 examples. CPs and EPs were transformed using min-max scaling. PCs were transformed using the Yeo-Johnson method due to extreme values. Lastly, ST features were cyclically encoded.

\subsection{Problem Formulation}\label{sec:prob_defn}
The problem is formulated as a MDP defined by the tuple $\langle \mathcal{S}, \mathcal{A},T,R,\gamma, D\rangle$, where $\mathcal{S}$ is the state space, $\mathcal{A}$ is the action space, ${T}: \mathcal{S}\times \mathcal{A} \times \mathcal{S} \rightarrow \mathbb{R}^{|\mathcal{S}|}$ is the transition function from a state $s$ to $s'$ after taking action $a$, $R: \mathcal{S}\times \mathcal{A} \rightarrow \mathbb{R}$ is the reward function, $\gamma \in (0,1]$ is the discount factor, and $D$ is the initial state distribution. The goal is to find a policy $\pi_{\theta} : S \xrightarrow{} \mathcal{A}$ that maximises expected cumulative reward, defined as $\mathbb{E}\left[\sum_{t=0}^H \gamma^t r(s_t,a_t) | s_0 \sim D, a_t  \sim \pi_{\theta}(s_t) \right]$, where $H$ is the planning horizon.

\textbf{States:} A raw state $s^{raw}_t\in\mathcal{S}$ at hour $t$ is composed of PCs observed between $t-1$ and $t$, EPs, and ST features. $s^{raw}_t$ is compressed into $s_t$  using an auto-encoder $g_{\psi} : S \rightarrow Z$, $S \in \mathbb{R}^{410}$, $Z \in \mathbb{R}^{50}$. 

\textbf{Actions:} The action space $\mathcal{A}$ is combinatorial, where each CP $i$ may take between 2 and 13 different discrete values represented by 
$
\mathit{CP}_i \equiv \{\mathit{CP}_{i,1}, \mathit{CP}_{i,2},$ $\ldots,  \mathit{CP}_{i,n_{\mathit{CP}_i}}\},
$
giving a total of $|\mathcal{A}|=\prod_{i=1}^N n_{\mathit{CP}_i} \approx 4.6\times10^{18}$ distinct actions for all 19 CPs. We train the policy to generate an action 
$
    a_t = (a_{1,t},a_{2,t},\ldots,a_{N,t}) \in \mathcal{A} 
$
at the beginning of hour $t$ given network state $s_t$ as input. Note that $a_{i,t} \in \mathit{CP}_i$ is a discrete value from the set $\mathit{CP}_i$.

\textbf{Reward:} Defined as the value of throughput KPI observed on a cell between $t$ and $t+1$. 

\textbf{Planning Horizon:} Single-step episodes are considered, in which an action is executed at time $t$, the reward is given at time $t+1$ and the episode terminates.


\subsection{Model-based Reinforcement Learning}

Since we are optimising over a single-step horizon, the learned dynamics function is limited to a one-step reward prediction model. This is defined as a probabilistic model $f_{\phi}(r_{t+1} | s_t, a_t) = Pr(r_{t+1} | s_t, a_t; \phi)$ that outputs the conditional distribution of the reward given the current state and action.
Learning a reward model is a task of fitting an approximation model to the true reward function given the training dataset $D_{train}=\{(s_i = g_{\psi}(s^{raw}_i), a_i=CP_i, r_i=TP_i)\}_{i=1}^M$ collected from the real network, where $g_{\psi} : S \rightarrow Z$, $S \in \mathbb{R}^{410}$, $Z \in \mathbb{R}^{50}$ is a feature extractor in the form of an auto-encoder neural network trained to reconstruct the raw state using dataset $\{(s_i^{raw})\}_{i=1}^M$. The motivation of pre-processing of raw state using an auto-encoder is to produce compact control polices for TP optimisation. Algorithm~\ref{alg:mbrl} presents the pseudo-code for model-based RL.

\begin{SCfigure}
  \centering
  \includegraphics[width=0.4\columnwidth]{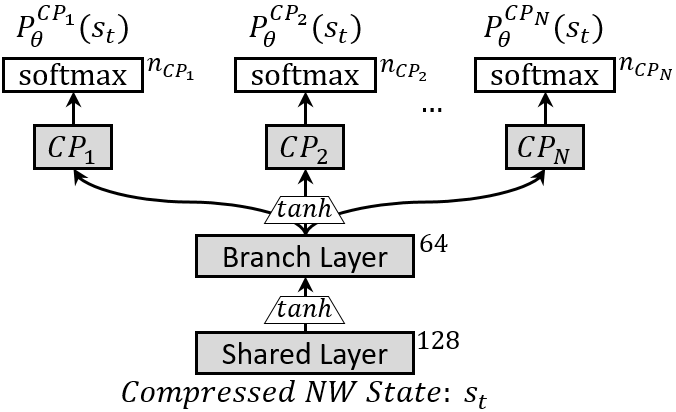}
  \caption{Architecture of the policy neural network $\pi_{\theta}(s|a)$, which is composed of $N$ sub-policies $\pi_{\theta}^i(a_i|s)$ each represented by a separate output head. Output heads share an embedding of the state computed
by two hidden layers. }
  \label{fig:rr_arch}
\end{SCfigure}

The architecture of the policy network $\pi_{\theta}(a | s)$ is shown in Figure~\ref{fig:rr_arch}. In our setting, the action components corresponding to CPs are chosen independently from each other. Given $N$ CPs, we decompose policy $\pi_{\theta}$ into $N$ sub-policies $\pi_{\theta}^i(a_i|s)$ such that  $\pi_{\theta}(a | s)= \prod_{i=1}^{N} \pi_{\theta}(a_i=CP_i | s)$. Decomposition of the policy network is achieved by branching $N$ output heads from a shared embedding of the state. The embedding is computed by two layers, where the first layer takes the compressed state $s_t$ as input. The $i^{th}$ output head returns a probability over the $i^{th}$ action component corresponding to $CP_i$. 

Proximal policy optimisation (PPO) \cite{schulman2017proximal} is employed to train $\pi_{\theta}(a | s)$. In step $k$, the policy update is given by:
\begin{equation*}
\centering
\theta_{k+1} = {\arg\max}_{\theta} \,\, \mathbb{E}_{s,a \sim \pi_ {\theta_{k}} } [{\cal L}_{PPO}(s,a, \theta_k, \theta)].
\end{equation*}
where the loss function is defined as:
\begin{equation}
\label{e:ppoloss}
\centering
\mathcal{L}_{PPO}(s,a, \theta_k, \theta) = \min \left( \frac{\pi_{\theta}(a | s)}{\pi_{\theta_{k}}(a | s)}A^{\pi_{\theta_{k}}}(s, a),\,\,\,\,g(\epsilon, A^{\pi_{\theta_{k}}}(s, a))   \right).
\end{equation}

\noindent The advantage $A^{\pi_{\theta_{k}}}(s, a)$ is clipped using function $g(\cdot)$ to prevent large changes in the policy parameters: $g(\epsilon, A) = (1 + \epsilon)A$, if $A \geq 0$; otherwise $g(\epsilon, A) = (1 - \epsilon)A$.
Advantages are estimated by subtracting the mean reward in a mini-batch $\mathcal{B}$ from the current reward:
\begin{equation}
    A^{\pi_{\theta_k}}(s,a) = r(s,a) - \frac{1}{|\mathcal{B}|}\sum_{i\in \mathcal{B}} r(s_i,a_i).
    \label{eqn:adv_reward}
\end{equation}

\vspace{-0.5cm}
\begin{algorithm}[h!]
\scriptsize
\caption{Model-based Reinforcement Learning}\label{alg:mbrl}
\begin{algorithmic}[1]
\State Collect dataset $D_{init}=\{(s_i^{raw},a_i=\mathit{CP}_i,r_i=\mathit{TP}_i)\}_{i=1}^M$ from real network.
\State Train feature extractor $g_{\psi}$ on dataset $\{(s_i^{raw})\}_{i=1}^M$
\State Process $D_{init}$ into dataset $D_{train} = \{(s_i=g_\psi(s_i^{raw}),a_i=\mathit{CP}_i, r_i=\mathit{TP}_i)\}_{i=1}^M$
\State Train reward model $f_{\phi}$ on $D_{train}$
\State Initialise control policy $\pi_{\theta}$
\For {\texttt{E} epochs}
  \State shuffle $D_{train}$ and partition it into mini-batches.
  \For {number of mini-batches}
        \State initialise training set $D_{RL}$
        \For {each example index \texttt{i} of a mini-batch}
            \State $\hat{a}_i$ $\sim$ $\pi_{\theta}(s_i)$ 
            \State $\hat{r}_i$ = $f_{\theta}(s_i, \hat{a}_i)$
            \State compute advantage $\hat{A}_i$ via Equation~\ref{eqn:adv_reward}
            \State add $(s_i, \hat{a}_i, \hat{A}_i)$ to $D_{RL}$
        \EndFor
        \State update $\pi_{\theta}$ based on $D_{RL}$ using the Adam optimiser, given loss in Equation \ref{e:ppoloss}
  \EndFor
\EndFor
\end{algorithmic}
\end{algorithm}

\vspace{-0.30cm}


\subsection{Compressing High Dimensional Network State}\label{sub_sec:NW_state_compression}

A total of 383 performance counters (PCs) are combined with 9 engineering parameters (EPs), and 18 spatio-temporal features (ST) to form the raw network state $s_t^{raw} \in \mathbb{R}^{410}$. An under-complete auto-encoder neural network $g_{\psi} : S \rightarrow Z$, $S \in \mathbb{R}^{410}$, $Z \in \mathbb{Z}^{50}$  
is trained to compress the raw network state. 

Compressed states produced by the encoder $g_\psi^\mathit{encoder}(s_t^{\mathit{raw}})\rightarrow s_t$ are passed to the decoder, which reconstructs the input: $g_\psi^\mathit{decoder}(s_t)\rightarrow \hat{s}_t^{\mathit{raw}}$. A regulariser $g_\psi^\mathit{reg}$ predicts the throughput at $t+1$ given $s_t$. Incorporating the regulariser was found to improve predictive accuracy of the reward model and policy gain. Loss is defined as the mean square error between the input state and reconstructed state:
\begin{equation*}
    \mathcal{L}(s^{\mathit{raw}},r,\psi) = (\lambda_\mathit{AE})\left(\frac{1}{|S|}\sum_{i\in S} \left(s^{\mathit{raw}}[i]-\hat{s}^{\mathit{raw}}[i]\right)^2\right) + (1-\lambda_\mathit{AE})(r-\hat{r})^2,
\end{equation*}
where $\hat{s}^{\mathit{raw}}$ is the reconstructed state produced by the decoder at time $t$, $\hat{r}$ is predicted reward (throughput) at $t+1$ given by the regulariser, and $\lambda_\mathit{AE} \in\left[0,1\right]$ weights contribution of the decoder and regulariser. Subscripts $t$ are dropped for clarity.

\vspace{-0.5cm}
\begin{figure}[!htb]
\begin{subfigure}{.58\textwidth}
    \centering
    \includegraphics[width= 1\linewidth]{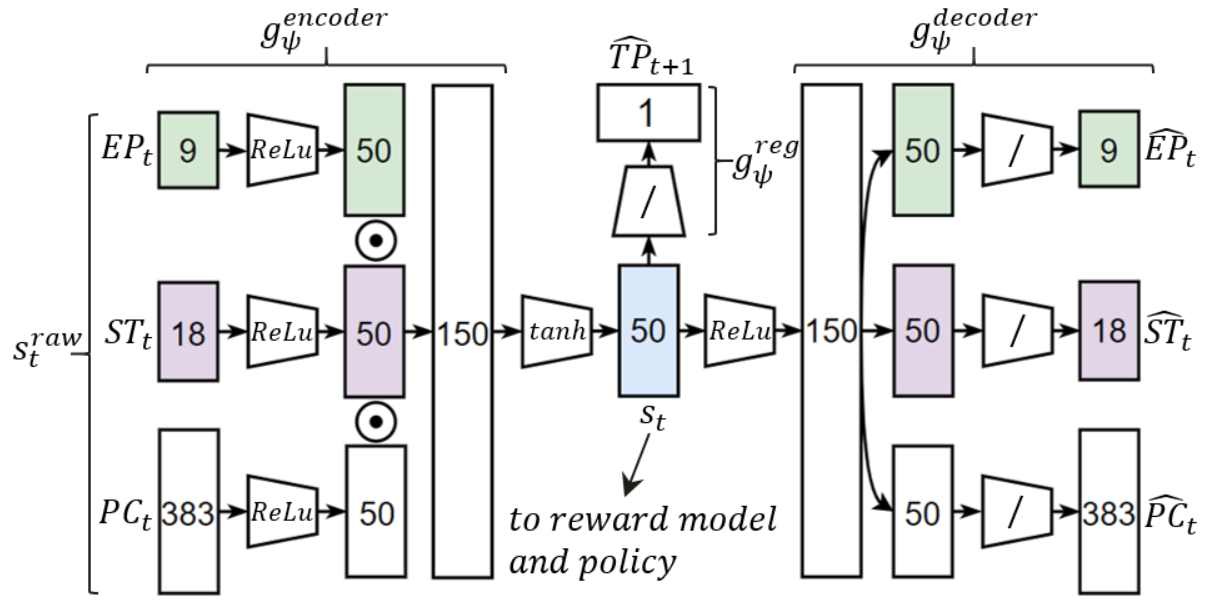}
    \caption{Network State Compressor.}
        \label{fig:AE_arch}
\end{subfigure}
\hspace{0.15cm}
\begin{subfigure}{.40\textwidth}
    \centering
    \includegraphics[width= 1\linewidth]{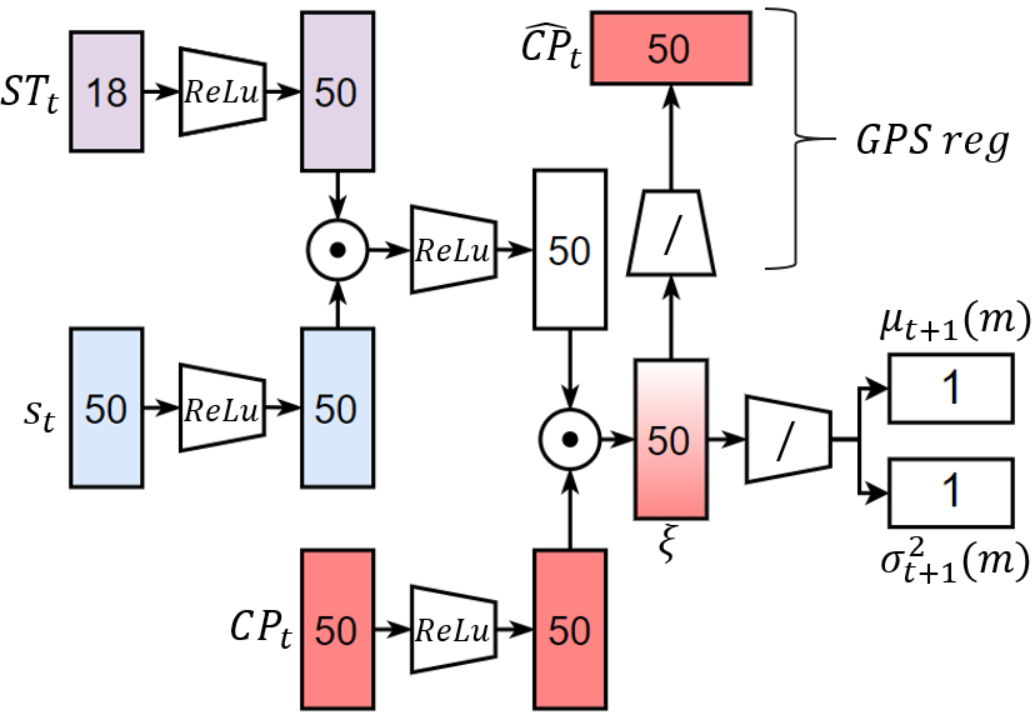}
    \caption{Reward Model.}
    \label{fig:SM_arch}
\end{subfigure}
 \caption{High dimensional raw network state $s^{raw}_t \in \mathbb{R}^{410}$ at hour $t$ is compressed using a modular auto-encoder to $s_t \in \mathbb{R}^{50}$. The reward model is an ensemble of probabilistic neural networks. A single component model $m$ is displayed.}
 \label{fig:NN_archs}
\end{figure}


\vspace{-0.5cm}

\subsection{Reward Model}\label{sub_sec:surrogate_model_design}
The reward model $f_{\phi} : S \times A \rightarrow \mathbb{R}$ is based on the dynamics model described by Chua et al.\,\,\cite{chua2018deep} termed as Probabilistic Ensemble with Trajectory Sampling. Predicted reward is given by a probabilistic ensemble consisting of 10 component models, each trained with different seeds and bootstrap samples of $D_{train}$ collected from the real network. Figure \ref{fig:SM_arch} displays the architecture of a single component model.  Each component model $f_\phi^i$ parameterises a Gaussian distribution with diagonal covariance: $P_\phi(r_{t+1}|s_t,\mathit{ST}_t,a_t)=\mathcal{N}(\mu_\phi(s_t,\mathit{ST}_t,a_t), \sigma_\phi(s_t,\mathit{ST}_t,a_t)^2)$, from which the predicted reward (throughput) at $t+1$ can be sampled. Inputs at hour $t$ include: the compressed network state $s_t$, current time and location of the cell $\mathit{ST}_t$, and actions produced by the policy $a_t=\mathit{CP}_t$.



We employ a causal inference-based regulariser to improve the generalisation of the reward model similar to the work of \cite{saini2019multiple}. Generalised Propensity Score (GPS) \cite{shi2019adapting} predicts the action $a_t=CP_t$ from a representation $\xi$ of $a_t$, compressed state $s_t$, and $ST_t$. The composite loss function contains terms for the negative log likelihood and GPS regulariser:
\begin{equation*}
    \mathcal{L}(s, a, \phi)
    = \lambda_\mathit{RM}\left(\frac{\left({\mathit{r}} - \mu_{\phi}(s,a)\right)^2}{\sigma_{\phi}^2(s,a)}+\log{\sigma_{\phi}^2(s,a)}\right)+\left(1-\lambda_\mathit{RM}\right)\mathrm{KL}\left(\mathit{a}  \| \widehat{\mathit{a}}\right),
\end{equation*}
where $r$ is the ground truth reward (throughput), $\mu_{\phi}(s,a)$ and $\sigma_{\phi}^2(s,a)$ are the predicted mean and variance, and $\lambda_\mathit{RM}\in\left[0,1\right]$ balances the contribution of the negative log likelihood and GPS regularisation terms. Subscripts $t$ are dropped for clarify. The overall reward model architecture is shown in Figure \ref{fig:SM_arch}.

Importantly, the probabilistic ensemble, consisting of component models $m\in\mathcal{M}$, quantifies the uncertainty associated with the predicted reward. Total uncertainty is decomposed into an epistemic term reflecting model uncertainty due to limited data, and an aleatoric term which represents irreducible noise:
\begin{equation*}
\sigma^2_\mathit{epistemic}(s,a) = \frac{1}{|{\mathcal{M}}|}\sum_{m\in\mathcal{M}}\left(\sigma^2_{\phi,m}(s,a)+\mu^2_{\phi,m}(s,a)\right) - \left(\frac{1}{|{\mathcal{M}}|}\sum_{m\in\mathcal{M}}\mu_{\phi,m}(s,a)\right)^2,  
\end{equation*}
\begin{equation*}
\sigma^2_\mathit{aleatoric}(x) = \frac{1}{|{\mathcal{M}}|}\sum_{m\in\mathcal{M}}\sigma^2_{\phi,m}(s,a).
\end{equation*}


\vspace{-0.95cm}
\begin{figure}[!htb]
\centering
\includegraphics[width=0.9\linewidth]{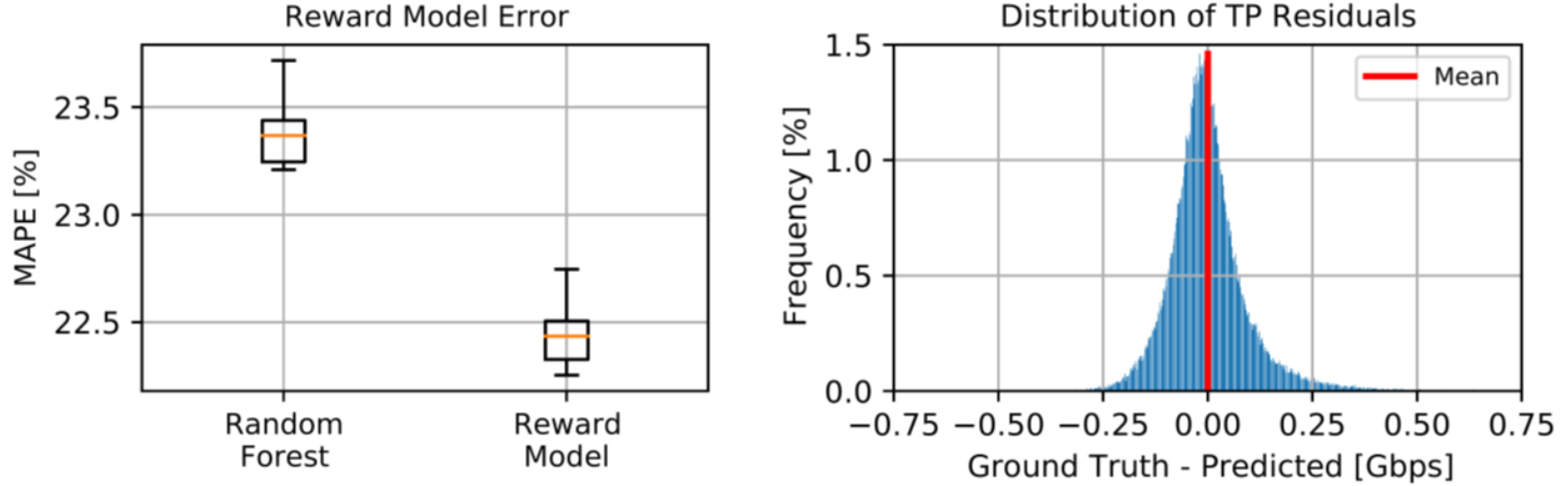}
 \caption{Analysis of the reward model.}
 \label{fig:SM_error_analysis}
\end{figure}

\vspace{-0.95cm}

\subsubsection{Reward Model Analysis:}
Experiments were carried out to confirm that the reward  model is an accurate simulator of the network environment dynamics. The neural network ensemble was bench-marked against a random forest containing 500 trees. Results from 10 fold cross-validation are displayed in Figure \ref{fig:SM_error_analysis}. Two conclusions are drawn from the plots. Firstly, the lowest error is achieved by the reward  model in all folds. Secondly, error residuals are symmetric around zero, confirming that throughput is neither systematically overestimated nor underestimated.

\subsection{Continual Reinforcement Learning}\label{sec:training_policies}



Continual RL is formulated as a sequence of MDPs $MDP(1), \ldots, MDP(L)$ where $\mathrm{MDP}(l)=\langle \mathcal{S}(l), \mathcal{A}(l),T(l),$ $R(l),\gamma, D(l)\rangle$~\cite{Khetarpal2020TowardsCR}, $l \in \{1, 2,\ldots, L\}$. In our case, each action space $\mathcal{A}(l)$ is a subset of the next: $\mathcal{A}(1) \subset \mathcal{A}(2) \subset \ldots \subset \mathcal{A}(L)$. Therefore each consecutive MDP in the sequence represents a new task with an expanding action space. 




We employ the Progress-\&-Compress (P\&C)  framework \cite{schwarz2018progress} for continual RL. P\&C enables the reuse of past information through layer-wise adaptors to a knowledge base. The policy consists of two columns: the active column acquires experience on a new task, and the knowledge base accumulates experience acquired over all previous tasks. The active column is trained during the \textbf{progress phase}. The knowledge base is then updated during the \textbf{compress phase}, in order to subsume  new knowledge acquired by the active column. Catastrophic forgetting of the knowledge base is prevented through a regularisation technique described below.
\vspace{-0.55cm}
\begin{figure}[!h]
  \centering
  \includegraphics[width=\columnwidth]{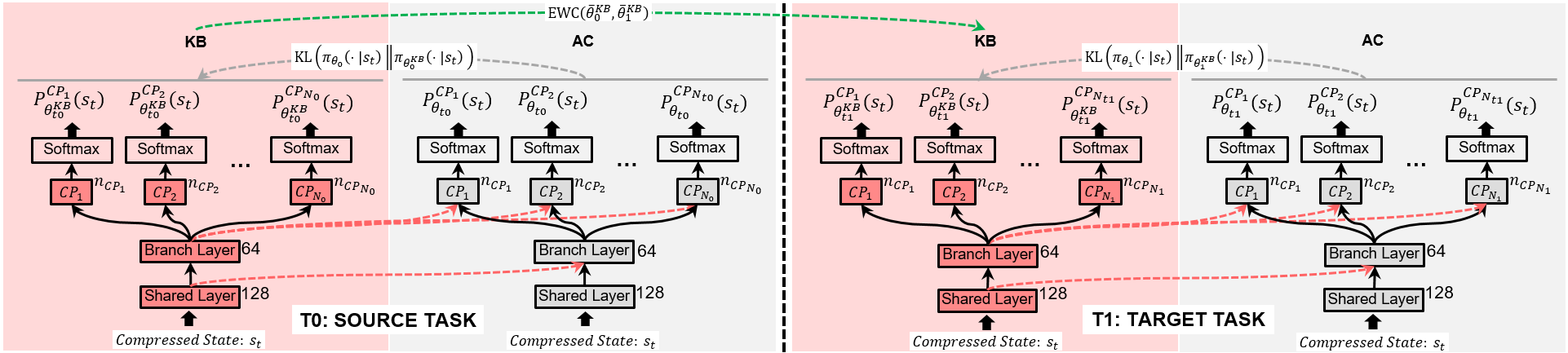}
  \caption{Training policies using progress and compress framework. Two hidden layers, called as \textit{shared} and \textit{branch}, are dense layers with $128$ and $64$ neurons, respectively, with $\mathrm{tanh}$ non-linearity.}
  \label{fig:pc_arch}
\end{figure}

\textbf{\textit{Progress Phase}:} When a new task is presented, the parameters of the knowledge base are frozen -- only those parameters in the active column are optimised. Layer-wise connections between the knowledge base and the active column allow  reuse of representations produced by the knowledge base, thus enabling positive transfer from previously learned tasks. Lateral adaptors are implemented as multi-layer perceptrons. The $j$-th layer of the active column is defined as follows:
\begin{equation*}
    h_j = \sigma\left(W_j h_{j-1} + \alpha_j \odot U_j \sigma\left(V_j h_{j-1}^{KB} + b_j^{KB}\right) + b_j\right),
\end{equation*}
where $\alpha_j$ is a trainable vector initialised by sampling from ${\cal U}(0,0.1)$ with size equal to number of units in layer $j$, and $W_j,U_j,V_j$ are weight matrices. The active column is trained using Algorithm~\ref{alg:mbrl}.

\textbf{\textit{Compress Phase}:} The compress phase distills the learned active column into the knowledge base. At this stage, the parameters of the active column are frozen. The compression optimisation objective is a distillation loss with the Elastic Weight Consolidation (EWC) penalty. EWC protects the knowledge base against catastrophic forgetting, such that all previously learned skills are maintained:
\begin{equation*}
    {\cal L}_{KB}\left(\theta_{T_t}^{KB}\right) = \mathbb{E}\left[\mathrm{KL}(\pi_{\theta_{T_t}}(\cdot|s) \| \pi_{\theta_{T_t}^{KB}}(\cdot|s))\right] + \mathrm{EWC}\left(\theta_{T_t}^{KB},\theta_{T_s}^{KB}\right)
\end{equation*}

\noindent where $\pi_{\theta_{T_t}}$ is the policy of active column after learning task $T_t$ while $\pi_{\theta_{T_t}^{KB}}$ is the knowledge base into which the active column is distilled.


The authors in \cite{schwarz2018progress} proposed online EWC to reduce the computational cost of calculating the Fisher information matrix:

\begin{equation*}
    \mathrm{EWC}(\theta_{T_t}^{KB},\theta_{T_s}^{KB}) = \frac{1}{2}\sum_{i\in {\cal I}^{KB}} \gamma_{\mathit{Fisher}} \left(F^*_{T_s,i}(\theta^{KB}_{T_t,i} - \theta^{KB}_{T_s,i})^2\right)
\end{equation*}

\noindent where $F^*_{T_t} = \gamma_{\mathit{Fisher}}\left(F^*_{T_s}\right) + \frac{1}{|\mathcal{B}|}\sum_{(s,a)\in \mathcal{S}\times\mathcal{A}}\nabla\log\pi_{\theta_{T_t}^{KB}}(a|s)\left( \nabla\log\pi_{\theta_{T_t}^{KB}}(a|s)^{\intercal}\right)$, ${\cal I}^{KB}$ are layers in the knowledge base excluding the output layers, $\mathcal{B}$ is the set of examples in the mini-batch, and $\gamma_{\mathit{Fisher}}\in\mathbb{R}$ is a hyper-parameter.

\section{Experimental Evaluation}\label{sec:evaluation}
A number of experiments are designed to assess the sample-efficiency of continual RL in the wireless optimisation scenarios, where the action space $CP_{source}$ in a source task $T_s$ is expanded to a set $CP_{target}$ in the target task $T_t$, where $CP_{source} \subset CP_{target}$. We are contrasting P\&C against a baseline \emph{Reinitialise-and-Retrain} (R\&R) method, which initialises $\pi_{\theta}(s|a)$ and trains it solely on the dataset $D_{init}$ specific to the CP set of a task in Algorithm~\ref{alg:mbrl}.

\subsection{Wireless Network Optimisation Scenarios}\label{sec:task_definitions}

    
    



\begin{SCfigure}
  \centering
  \includegraphics[width=0.65\columnwidth]{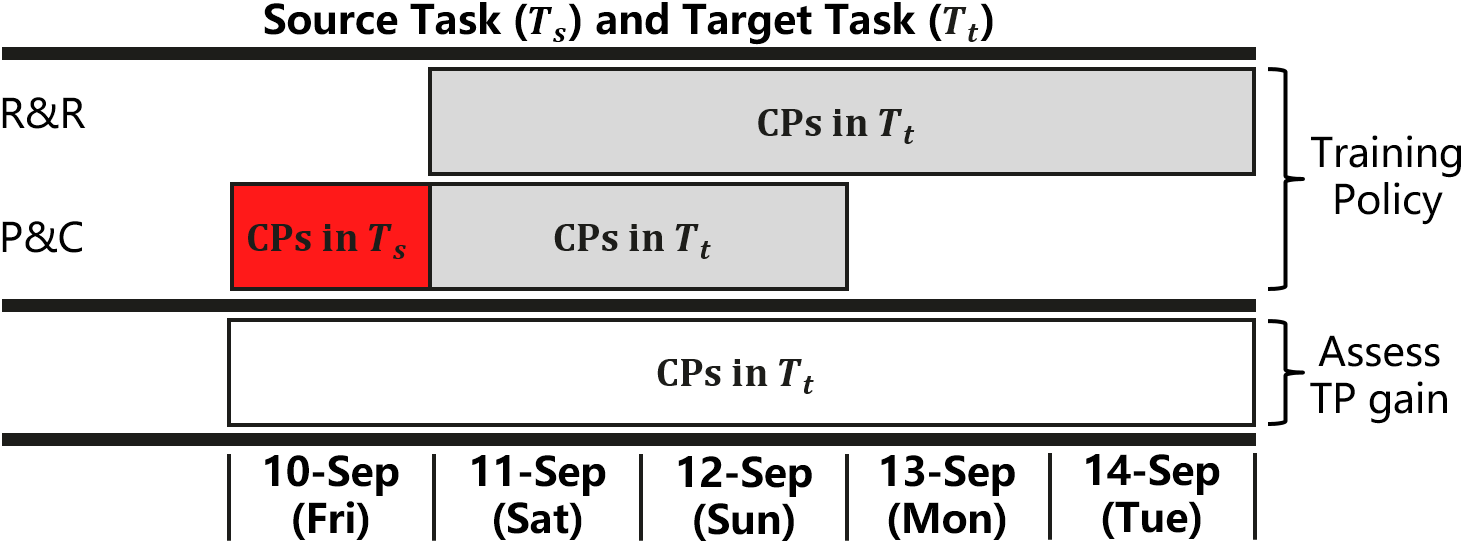}
  \caption{Policies trained via Reinitialise-\&-Retrain (R\&R) use four days of data from the target task $(T_t)$. Policies trained via Progress-\&-Compress (P\&C) use two days of data from $T_t$ (50\% less), and transfer knowledge from a source task $T_s$. 
  \label{fig:self_extension_scenarios}}
\end{SCfigure}

We defined three scenarios that demonstrate how domain knowledge of a network operator is used to optimise sets of CPs for a new network site in a stage-wise manner.

\textbf{Scenario 1:} Use of prior knowledge about the environment in which the cell is deployed (e.g. height of surrounding buildings, density of buildings) and the profile of services that are expected in this geographical area (e.g. ultra reliable low-latency communication). In this scenario, based on domain knowledge, the operator chooses initially a small subset of CPs known to have high impact to the optimisation objective KPIs. The small subset is then extended to the full set of CPs to leverage additional CP interactions.

\textbf{Scenario 2:} Use of prior knowledge about the potential of certain CPs to cause network disruptions or network performance degradation (i.e. exploring CPs related to energy saving sleep-modes of a base-station, or CPs that determine antenna configurations directly impacting coverage). In this scenario, network optimisation starts with a set of CPs known to have low impact on the optimisation objective KPIs in order to reduce said risks. The initial set is then gradually extended to the full set of CPs.

\textbf{Scenario 3:} A combination of considerations reflected in Scenarios 1 and 2.

\subsection{Policy Training Setup} \label{s:trainingsetup}
Five experiments were designed to assess policy deployment lead-time reduction in the three scenarios described above. The experimental setups are outlined in Table \ref{tab:expt_design}. In each experimental setup, policies trained using P\&C were initialised on a source task $T_s$ and updated on the target task $T_t$. Policies trained using R\&R were trained on the $T_t$ using twice the training data volume as P\&C (4 days as opposed to 2 days). 
\begin{table}[htbp]
\scriptsize
\caption{Experimental setups to assess policy deployment lead-time reduction after expanding the set of CPs from a source task ($T_s$) to target task ($T_t$).}
\setlength\extrarowheight{2.0pt} 
\begin{tabu} to 1\textwidth { | X[l] | X[l] |  X[l] | X[l]  |}
   \hline
   \textbf{Experiment} & \textbf{Scenario} & \textbf{CPs in $T_s$} & \textbf{CPs in $T_t$} \\
   \hline
   1 & 1 & 4 high-impact  & All 19 \\
   \hline    
    2 & 2 & 15 low-impact & All 19 \\
    \hline
    3 & 1 & 2 high-impact & 4 high-impact  \\
    \hline
    4 & 3 & 2 high-impact \& 7 low-impact & All 19  \\
    \hline
    5  & 2 & 7 low-impact & 15 low-impact  \\
    \hline
\end{tabu}
\label{tab:expt_design}
\end{table}

To ensure an unbiased evaluation, the dataset $D_{init}$ was partitioned into disjoint subsets as follows: 
\begin{itemize}
    \item 30\% of the examples were reserved to train a test system model with similar structure to the reward model.
    \item Of the 70\% remaining; 70\% of the examples were used to train the reward model, state compressor, and policies, and 30\% was used to assess TP gain as described in the following section.
\end{itemize}

\noindent The datasets were further structured as outlined in Figure \ref{fig:self_extension_scenarios} in order to evaluate P\&C under a 50\% reduction in data volume compared to R\&R.

\vspace{-0.5cm}
\begin{table}[htbp]
\scriptsize
\caption{Hyperparameters used in the experimental evaluation.}
\setlength\extrarowheight{2.0pt} 
\begin{tabu} to 1\textwidth { | X[l] | X[l] |  X[l] | X[l] | X[l] |}
   \hline
   \textbf{Hyperparameter} & \textbf{State Compressor} & \textbf{Reward Model} & \textbf{R\&R} & \textbf{P\&C}\\
   \hline
    Training Iterations & $\mathit{epochs}=100$ & $\mathit{epochs}=100$ & $\mathit{epochs}=100$ & $\mathit{steps}=1000$ \\
   \hline    
    Learning Rate & 0.00015 & 0.00015 & 0.0001 & 0.0001\\
    \hline
    Batch Size & 128 & 128 & 64 & 64 \\
    \hline
    $\gamma_{\mathit{Fisher}}$ & NA & NA & NA & 0.9 \\
    \hline
    $\lambda$  & $\lambda_{\mathit{AE}}=0.9$ & $\lambda_{\mathit{RM}}=0.9$ & NA & NA \\
    \hline
\end{tabu}
\label{tab:challenges}
\end{table}

\subsection{Performance Metrics}\label{sec:metrics}

Policy deployment is constrained by real-time, and its lead-time is approximately equivalent to the time required for the collection of dataset $D_{init}$ in Algorithm~\ref{alg:mbrl}.  Our target is to demonstrate (i) 50\% reduction in the amount of data required by P\&C relative to R\&R, and (ii) positive TP gain. TP gain in hour $t$ is defined as:

\begin{equation}
    \mathit{TP\,Gain}(t)\,\,[\%] = \frac{\sum_{c\in\mathcal{C}} \widehat{TP}_{c,t}^{P\&C} - \sum_{c\in\mathcal{C}} \widehat{TP}_{c,t}^{R\&R}}{\sum_{c\in\mathcal{C}} \widehat{TP}_{c,t}^{R\&R}}\times 100\,\%, 
\label{eq:TP_GAIN}
\end{equation}

\noindent where $\mathcal{C}$ is the set of all cells, $\widehat{TP}_{c,t}^{P\&C}$ is the predicted TP when cell $c\in\mathcal{C}$ executes CPs given by the P\&C policy and $\widehat{TP}_{c,t}^{R\&R}$ is the predicted TP when cell $c$ executes CPs given by the R\&R policy. Predictions are given by the test system model. TP gain is computed in the test set as described in Section~\ref{s:trainingsetup}.

\section{Results}\label{sec:results}

Results are reported based on 20 independent runs of each experiment in Table~\ref{tab:expt_design}. 
\subsubsection{Training performance:}  
\begin{figure}[!h]
  \centering
  \includegraphics[width=1.0\columnwidth]{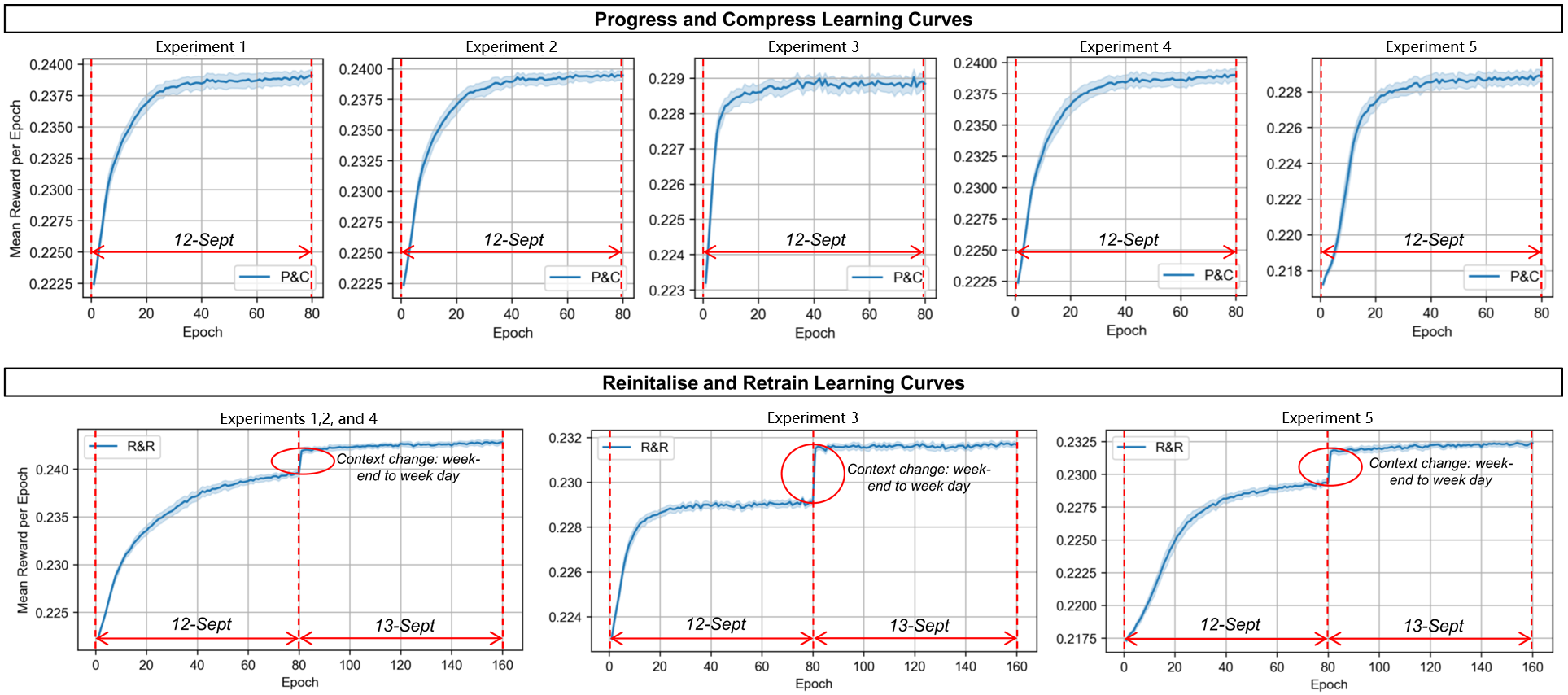}
  \caption{Learning curves. Blue shading shows standard deviation over 20 independent runs.\label{fig:learning_curves}} 
\end{figure}

Figure \ref{fig:learning_curves} shows the learning curves of P\&C and R\&R. First, we observe that the runs of experiments for Scenario 1 that start with high-impact CPs in the source task converge faster than those runs of Scenario 2, which start with low-impact CPs is source task. A context change occurs between September 12\textsuperscript{th} and 13\textsuperscript{th} when training R\&R policies incrementally. The context change is associated with switching from the weekend to a week day. 
\vspace{-0.55cm}

The jump in mean reward per epoch on September 13\textsuperscript{th} is caused by a change in the underlying user demand. This change is due to more users occupying hotspots such as downtown business districts, transport hubs, etc. during weekdays. These areas are served by high capacity cells, resulting in a higher overall throughput. 





\begin{figure}[!h]
  \centering \includegraphics[width=0.65\columnwidth]{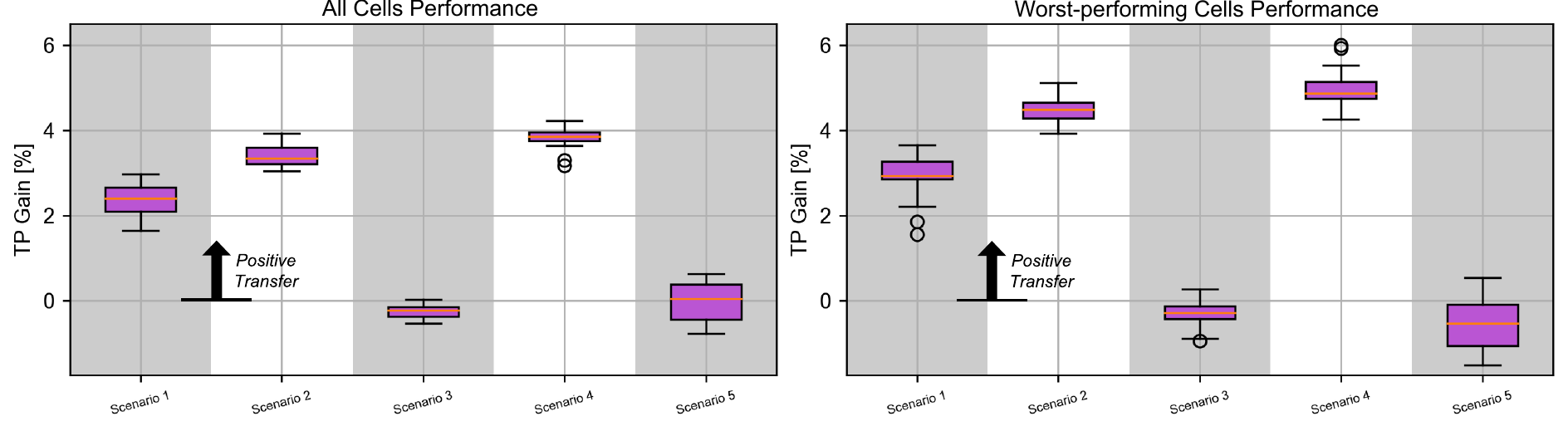}
  \caption{Assessment of TP gain.  \label{fig:tp_gain}} 
\end{figure}

\begin{figure}[!h]
  \centering \includegraphics[width=0.85\columnwidth]{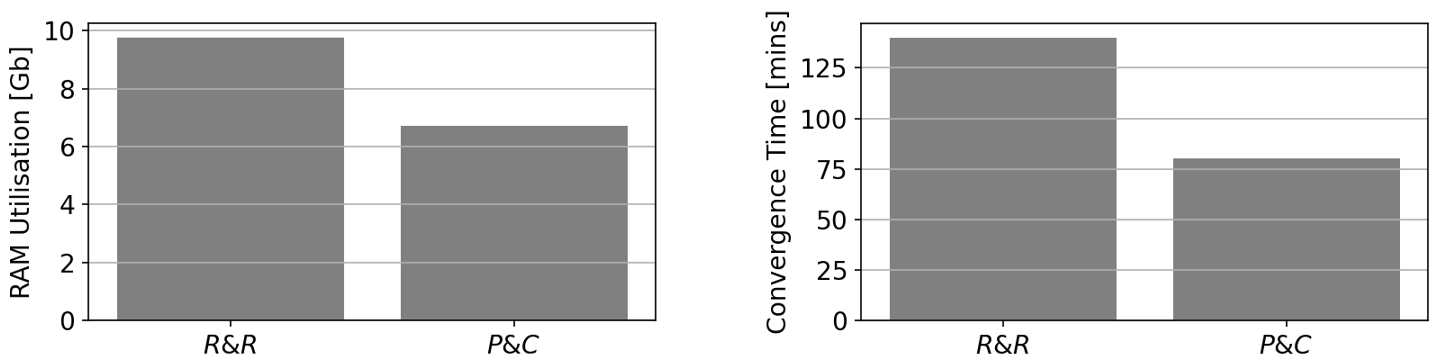}
  \caption{Training resource footprint.  \label{fig:resfoot}} 
\end{figure}


\subsubsection{Assessment of TP gain:} Figure \ref{fig:tp_gain} illustrates the distribution of median TP gain over 20 independent runs calculated at hourly intervals as defined in Equation \ref{eq:TP_GAIN}. The median of TP over 20 independent runs for the P\&C and R\&R methods is given by $TP_{c,t}^{X} = \mathrm{median}(TP_{c,t,1}^{X},TP_{c,t,2}^{X},\ldots,TP_{c,t,20}^{X}), \quad X\in \{P\&C, R\&R\}$, where $TP_{c,t,run}^{X}$ is the throughput for cell $c$ at time $t$ for $run = 1,2,\ldots,20$.

Results displayed in Figure~\ref{fig:tp_gain} confirm that P\&C achieves a positive median TP gain along with a data reduction of 50\% relative to the baseline R\&R policy in experiments 1, 2, 4, and 5. The highest median TP gain is attained in experiment 4 due to the presence of both high-impact and low-impact CPs in the source task. P\&C is most effective in experiments 1, 2, and 4 with all 19 CPs in the target task. We observe smaller positive median TP gain in experiment 5 because of the low impact CPs in both source and target tasks, while a negative median TP gain is observed in experiment 3 due to few low-impact CPs in the source task. This result suggests that the effectiveness of continual RL as a sequence of MDPs with an expanding combinatorial action space is mostly affected by the strength of the causal effect of each consecutive action subset.


\subsubsection{Training resource cost:}

The amount of training RAM and convergence wall time are illustrated in Figure~\ref{fig:resfoot}. $P\&C$ results in a reduction of approx. 31\% in memory footprint compared to $R\&R$ baseline, amounting to a total of $6.7$ GB. This is crucial for training in resource-constrained hardware, which is typical for on-premise deployments in  telecommunication networks. Average training wall time for $P\&C$ is measured at 80 minutes compared to 140 minutes for $R\&R$. Reduced training time allows us to allocate more resources for hyper-parameter tuning and model selection, and enables more frequent model updates in highly-dynamic wireless networks.

\subsubsection{Inference time:} Inference for setting configuration parameters reduces to a forward pass of the compact neural net policy, and it was measured at 80 milliseconds of wall-time for 20,000 cells.

\section{Conclusions and Future Work}
\label{sec:conclusions}

Learning optimisation policies from scratch in order to assess the performance of different subsets of configuration parameters on the objective KPI is costly is terms of time required for data collection, particularly for unknown wireless network environments with exploration challenges (i.e. safety constraints, limited budget of interactions with the wireless network, high dimensional state/action spaces, high noise levels). The contribution of our work is the application of Continual Reinforcement Learning to a wireless network optimisation system where the number of decision variables gradually increase throughout its lifetime. By refraining from reinitialising-and-retraining the policy, we achieve a two-fold reduction in end-to-end deployment time for each consecutive decision variable subset, which has direct implications in reducing operational cost and overall data collection requirement.



We defined three scenarios reflecting how domain knowledge of a network operator can be leveraged to handcraft a series of CP subsets to be optimised stage-wise. Five experiments based on these scenarios were conducted, and empirical results demonstrated a positive throughput gain of up to 4\% of P\&C continual RL method against the R\&R baseline that used double the amount of training data. 



The selection of CP subsets for optimisation policies deployed to new wireless network sites is traditionally an iterative trial-and-error process that predominantly relies on domain knowledge. Going forward, there are two approaches that will be investigated to further improve sample efficiency. The first approach is based on causal structure learning to automatically identify the CPs that exert the strongest causal effect on the optimisation objective KPI in a new environment, and hence complement and fine-tune operator's domain knowledge during selection of CP subsets. Of applicability here are methods that admit latent confounders when learning causal graphs, for example the works of~\cite{Colombo2012LearningHD,ogarrio16,Raghu2018ComparisonOS} and also the auto-causal tuning method of~\cite{9828518}. The second approach is based on policy warm-starting through the technology of \emph{sim-to-real transfer} in RL~\cite{9308468}, some examples of which include domain adaptation, imitation learning and policy distillation.



\bibliography{bibliography}

\begin{thebibliography}{10}
\providecommand{\url}[1]{\texttt{#1}}
\providecommand{\urlprefix}{URL }
\providecommand{\doi}[1]{https://doi.org/#1}

\bibitem{DBLP:journals/tccn/BaleviA19}
Balevi, E., Andrews, J.G.: Online antenna tuning in heterogeneous cellular
  networks with deep reinforcement learning. {IEEE} Trans. Cogn. Commun. Netw.
  \textbf{5}(4),  1113--1124 (2019). \doi{10.1109/TCCN.2019.2933420},
  \url{https://doi.org/10.1109/TCCN.2019.2933420}

\bibitem{9828518}
Biza, K., Tsamardinos, I., Triantafillou, S.: Out-of-sample tuning for causal
  discovery. IEEE Transactions on Neural Networks and Learning Systems pp.
  1--11 (2022). \doi{10.1109/TNNLS.2022.3185842}

\bibitem{DBLP:conf/blackseecom/BotheMFI20}
Bothe, S., Masood, U., Farooq, H., Imran, A.: Neuromorphic {AI} empowered root
  cause analysis of faults in emerging networks. In: {IEEE} International Black
  Sea Conference on Communications and Networking, BlackSeaCom 2020, Odessa,
  Ukraine, May 26-29, 2020. pp.~1--6. {IEEE} (2020).
  \doi{10.1109/BlackSeaCom48709.2020.9235002},
  \url{https://doi.org/10.1109/BlackSeaCom48709.2020.9235002}

\bibitem{ericsson}
Bouton, M., Farooq, H., Forgeat, J., Bothe, S., Shirazipour, M., Karlsson, P.:
  Coordinated reinforcement learning for optimizing mobile networks (09 2021)

\bibitem{Calabrese2018LearningRR}
Calabrese, F.D., Wang, L., Ghadimi, E., Peters, G., Hanzo, L., Soldati, P.:
  Learning radio resource management in rans: Framework, opportunities, and
  challenges. IEEE Communications Magazine  \textbf{56},  138--145 (2018)

\bibitem{chua2018deep}
Chua, K., Calandra, R., McAllister, R., Levine, S.: Deep reinforcement learning
  in a handful of trials using probabilistic dynamics models. Advances in
  neural information processing systems  \textbf{31} (2018)

\bibitem{Colombo2012LearningHD}
Colombo, D., Maathuis, M.H., Kalisch, M., Richardson, T.S.: Learning
  high-dimensional directed acyclic graphs with latent and selection variables.
  The Annals of Statistics  \textbf{40}(1),  294--321 (2012)

\bibitem{10.1007/s11277-016-3849-9}
Dandanov, N., Al-Shatri, H., Klein, A., Poulkov, V.: Dynamic self-optimization
  of the antenna tilt for best trade-off between coverage and capacity in
  mobile networks. Wirel. Pers. Commun.  \textbf{92}(1),  251–278 (jan 2017).
  \doi{10.1007/s11277-016-3849-9},
  \url{https://doi.org/10.1007/s11277-016-3849-9}

\bibitem{devin2017learning}
Devin, C., Gupta, A., Darrell, T., Abbeel, P., Levine, S.: Learning modular
  neural network policies for multi-task and multi-robot transfer. In: 2017
  IEEE international conference on robotics and automation (ICRA). pp.
  2169--2176. IEEE (2017)

\bibitem{conf/vtc/EckhardtKG11}
Eckhardt, H., Klein, S., Gruber, M.: Vertical antenna tilt optimization for lte
  base stations. In: VTC Spring. pp.~1--5. IEEE (2011),
  \url{http://dblp.uni-trier.de/db/conf/vtc/vtc2011s.html#EckhardtKG11}

\bibitem{DBLP:conf/pimrc/EisenblatterG08}
Eisenbl{\"{a}}tter, A., Geerdes, H.: Capacity optimization for {UMTS:} bounds
  and benchmarks for interference reduction. In: Proceedings of the {IEEE} 19th
  International Symposium on Personal, Indoor and Mobile Radio Communications,
  {PIMRC} 2008, 15-18 September 2008, Cannes, French Riviera, France. pp.~1--6.
  {IEEE} (2008). \doi{10.1109/PIMRC.2008.4699919},
  \url{https://doi.org/10.1109/PIMRC.2008.4699919}

\bibitem{isele2018selective}
Isele, D., Cosgun, A.: Selective experience replay for lifelong learning. In:
  Proceedings of the AAAI Conference on Artificial Intelligence. vol.~32 (2018)

\bibitem{Khetarpal2020TowardsCR}
Khetarpal, K., Riemer, M., Rish, I., Precup, D.: Towards continual
  reinforcement learning: A review and perspectives. J. Artif. Intell. Res.
  \textbf{75},  1401--1476 (2020)

\bibitem{kirkpatrick2017overcoming}
Kirkpatrick, J., Pascanu, R., Rabinowitz, N., Veness, J., Desjardins, G., Rusu,
  A.A., Milan, K., Quan, J., Ramalho, T., Grabska-Barwinska, A., et~al.:
  Overcoming catastrophic forgetting in neural networks. Proceedings of the
  national academy of sciences  \textbf{114}(13),  3521--3526 (2017)

\bibitem{li2017learning}
Li, Z., Hoiem, D.: Learning without forgetting. IEEE transactions on pattern
  analysis and machine intelligence  \textbf{40}(12),  2935--2947 (2017)

\bibitem{mendez2022modular}
Mendez, J.A., van Seijen, H., Eaton, E.: Modular lifelong reinforcement
  learning via neural composition. arXiv preprint arXiv:2207.00429  (2022)

\bibitem{8792117}
Nasir, Y.S., Guo, D.: Multi-agent deep reinforcement learning for dynamic power
  allocation in wireless networks. IEEE Journal on Selected Areas in
  Communications  \textbf{37}(10),  2239--2250 (2019).
  \doi{10.1109/JSAC.2019.2933973}

\bibitem{ogarrio16}
Ogarrio, J.M., Spirtes, P., Ramsey, J.: A hybrid causal search algorithm for
  latent variable models. In: Proceedings of the Eighth International
  Conference on Probabilistic Graphical Models. vol.~52, pp. 368--379. PMLR
  (06--09 Sep 2016)

\bibitem{10.1109/TNET.2013.2294965}
Partov, B., Leith, D.J., Razavi, R.: Utility fair optimization of antenna tilt
  angles in lte networks. IEEE/ACM Trans. Netw.  \textbf{23}(1),  175–185
  (feb 2015). \doi{10.1109/TNET.2013.2294965},
  \url{https://doi.org/10.1109/TNET.2013.2294965}

\bibitem{Raghu2018ComparisonOS}
Raghu, V.K., Ramsey, J., Morris, A., Manatakis, D., Spirtes, P., Chrysanthis,
  P.K., Glymour, C., Benos, P.: Comparison of strategies for scalable causal
  discovery of latent variable models from mixed data. International Journal of
  Data Science and Analytics  \textbf{6},  33 -- 45 (2018)

\bibitem{rolnick2019experience}
Rolnick, D., Ahuja, A., Schwarz, J., Lillicrap, T., Wayne, G.: Experience
  replay for continual learning. Advances in Neural Information Processing
  Systems  \textbf{32} (2019)

\bibitem{rusu2015policy}
Rusu, A.A., Colmenarejo, S.G., Gulcehre, C., Desjardins, G., Kirkpatrick, J.,
  Pascanu, R., Mnih, V., Kavukcuoglu, K., Hadsell, R.: Policy distillation.
  arXiv preprint arXiv:1511.06295  (2015)

\bibitem{rusu2016progressive}
Rusu, A.A., Rabinowitz, N.C., Desjardins, G., Soyer, H., Kirkpatrick, J.,
  Kavukcuoglu, K., Pascanu, R., Hadsell, R.: Progressive neural networks. arXiv
  preprint arXiv:1606.04671  (2016)

\bibitem{saini2019multiple}
Saini, S.K., Dhamnani, S., Ibrahim, A.A., Chavan, P.: Multiple {T}reatment
  {E}ffect {E}stimation using {D}eep {G}enerative {M}odel with {T}ask
  {E}mbedding. In: The World Wide Web Conference. pp. 1601--1611 (2019)

\bibitem{schulman2017proximal}
Schulman, J., Wolski, F., Dhariwal, P., Radford, A., Klimov, O.: Proximal
  policy optimization algorithms. arXiv preprint arXiv:1707.06347  (2017)

\bibitem{schwarz2018progress}
Schwarz, J., Czarnecki, W., Luketina, J., Grabska-Barwinska, A., Teh, Y.W.,
  Pascanu, R., Hadsell, R.: Progress \& {C}ompress: {A} scalable framework for
  continual learning. In: International Conference on Machine Learning. pp.
  4528--4537. PMLR (2018)

\bibitem{DBLP:journals/twc/ShafinCNHPZRL20}
Shafin, R.S.B., Chen, H., Nam, Y., Hur, S., Park, J., Zhang, J., Reed, J.H.,
  Liu, L.: Self-tuning sectorization: Deep reinforcement learning meets
  broadcast beam optimization. {IEEE} Trans. Wirel. Commun.  \textbf{19}(6),
  4038--4053 (2020). \doi{10.1109/TWC.2020.2979446},
  \url{https://doi.org/10.1109/TWC.2020.2979446}

\bibitem{shi2019adapting}
Shi, C., Blei, D., Veitch, V.: Adapting {N}eural {N}etworks for the
  {E}stimation of {T}reatment {E}ffects. Advances in neural information
  processing systems  \textbf{32} (2019)

\bibitem{traore2019discorl}
Traor{\'e}, R., Caselles-Dupr{\'e}, H., Lesort, T., Sun, T., Cai, G.,
  D{\'\i}az-Rodr{\'\i}guez, N., Filliat, D.: Discorl: Continual reinforcement
  learning via policy distillation. arXiv preprint arXiv:1907.05855  (2019)

\bibitem{DBLP:conf/vtc/VannellaJP20}
Vannella, F., Jeong, J., Prouti{\`{e}}re, A.: Off-policy learning for remote
  electrical tilt optimization. In: 92nd {IEEE} Vehicular Technology
  Conference, {VTC} Fall 2020, Victoria, BC, Canada, November 18 - December 16,
  2020. pp.~1--5. {IEEE} (2020). \doi{10.1109/VTC2020-Fall49728.2020.9348456},
  \url{https://doi.org/10.1109/VTC2020-Fall49728.2020.9348456}

\bibitem{yin2017knowledge}
Yin, H., Pan, S.J.: Knowledge transfer for deep reinforcement learning with
  hierarchical experience replay. In: Thirty-First AAAI conference on
  artificial intelligence (2017)

\bibitem{yoon2017lifelong}
Yoon, J., Yang, E., Lee, J., Hwang, S.J.: Lifelong learning with dynamically
  expandable networks. arXiv preprint arXiv:1708.01547  (2017)

\bibitem{9308468}
Zhao, W., Queralta, J.P., Westerlund, T.: Sim-to-real transfer in deep
  reinforcement learning for robotics: a survey. In: 2020 IEEE Symposium Series
  on Computational Intelligence (SSCI). pp. 737--744 (2020).
  \doi{10.1109/SSCI47803.2020.9308468}

\end{thebibliography}
\bibliographystyle{splncs04}
\end{document}